\documentclass[letterpaper, 10 pt, conference]{ieeeconf}  

\IEEEoverridecommandlockouts                              

\usepackage{amsmath} 
\usepackage{amssymb}  
\usepackage{booktabs}
\usepackage{graphicx}
\usepackage{cite}

\title{\LARGE \bf
FlipConcept: Tuning-Free Multi-Concept Personalization for Text-to-Image Generation
}

\author{Young-Beom Woo$^{1}$, Sun-Eung Kim$^{1}$, and Seong-Whan Lee$^{1}$
\thanks{{*}This research was supported by the Institute of Information \& Communications Technology Planning \& Evaluation (IITP) grant, funded by the Korea government (MSIT) (No. RS-2019-II190079 (Artificial Intelligence Graduate School Program (Korea University)), and No. IITP-2025-RS-2024-00436857 (Information Technology Research Center (ITRC)).}%
\thanks{$^{1}$Y.-B. Woo, S.-E. Kim, and S.-W. Lee are with the Department of Artificial Intelligence, Korea University, Anam-dong, Seongbukku, Seoul 02841, Korea. 
{\texttt{\small\{woo\_y\_b, se\_kim, and sw.lee\}@korea.ac.kr}}%
}}
\begin{document}

\maketitle
\thispagestyle{empty}
\pagestyle{empty}

\begin{figure*}[t]
\centering
\includegraphics[width=0.66\linewidth]{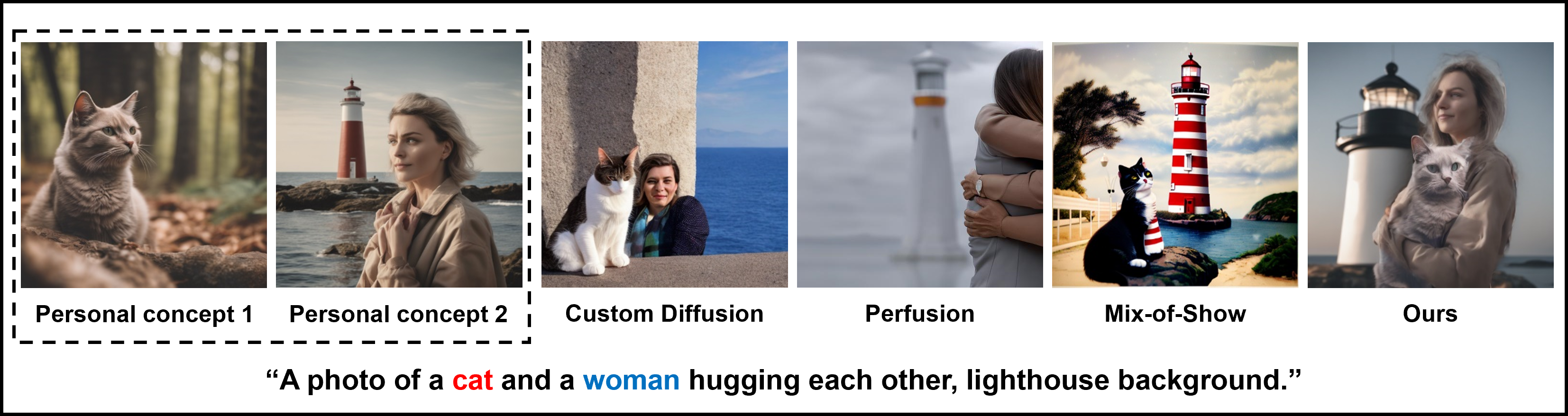}
  \caption{\textbf{Comparison of multi-subject personalization methods.} The first two columns illustrate the learned personalized concepts (the cat and the woman).
From left to right, we show the final generated results of Custom Diffusion, Perfusion, Mix-of-Show, and FlipConcept (Ours), all prompted with:
\textit{``A photo of a cat and a woman hugging each other, lighthouse background.''}
While existing approaches often cause some concepts to disappear, blend appearances, or distort structural details when dealing with multiple subjects, our method preserves each concept’s features and the overall scene context, resulting in coherent and high-quality outputs.}
\label{fig:short}
\end{figure*}

\begin{abstract}

Integrating multiple personalized concepts into a single image has recently gained attention in text-to-image (T2I) generation.
However, existing methods often suffer from performance degradation in complex scenes due to distortions in non-personalized regions and the need for additional fine-tuning, limiting their practicality. 
To address this issue, we propose FlipConcept, a novel approach that seamlessly integrates multiple personalized concepts into a single image without requiring additional tuning. We introduce guided appearance attention to enhance the visual fidelity of personalized concepts. Additionally, we introduce mask-guided noise mixing to protect non-personalized regions during concept integration. Lastly, we apply background dilution to minimize concept leakage, i.e., the undesired blending of personalized concepts with other objects in the image. In our experiments, we demonstrate that the proposed method, despite not requiring tuning, outperforms existing models in both single and multiple personalized concept inference. These results demonstrate the effectiveness and practicality of our approach for scalable, high-quality multi-concept personalization.

\end{abstract}

\section{Introduction}

Text-to-image (T2I) generation has made significant progress in synthesizing images based on user-provided personalized concepts.  
Users increasingly want to combine multiple personal concepts, such as a pet or a favorite object, into a single coherent image that captures a specific scene or memory.
This task remains challenging, as models must arrange concepts by the prompt while preserving visual identities and spatial relations.
Custom Diffusion~\cite{kumari2023multi}, a representative personalization approach, fine-tunes the cross-attention layers of a pre-trained diffusion model using multiple personalized images.  
Nevertheless, tuning-based methods~\cite{kumari2023multi,gal2022image,ruiz2023dreambooth,tewel2023key,gu2024mix,kwon2024concept} are prone to overfitting and struggle to generalize when combining multiple concepts.  
These models often suffer from concept interference or leakage, where the characteristics of one concept unintentionally affect another, leading to visual artifacts, reduced fidelity, or loss of prompt relevance.  
To address these issues, recent research has explored tuning-free approaches~\cite{alaluf2024cross, cao2023masactrl, nam2024dreammatcher}.

However, despite their remarkable achievements, existing methods still frequently fail to manage interactions among multiple objects effectively as shown in Fig.~\ref{fig:short}.
In particular, in complex scenarios, such as balancing multiple objects within an image, these methods often fail to maintain the relationships between concepts or handle non-personalized regions appropriately, leading to unintended concept mixing, distortion, or modification~\cite{lee2015motion}.

To address these challenges, we propose a novel method called \textbf{FlipConcept}, a modular inference stage pipeline that consistently generates images containing multiple personalized concepts using just a single reference image per concept, without requiring additional fine-tuning.
Unlike prior tuning-based approaches, FlipConcept does not alter model parameters, enabling enhanced scalability and practicality while maintaining both concept fidelity and spatial coherence.
FlipConcept is the first method to utilize Edit-Friendly DDPM Inversion~\cite{huberman2024edit} for multi-concept personalization, achieving tuning-free integration and superior concept fidelity compared to existing methods.
Our framework operates in two distinct stages.

In the first stage, we prepare the input data for the model.  
Specifically, we generate a background image and extract masks from it.  
Then, we perform Edit-Friendly DDPM Inversion~\cite{huberman2024edit} on the prepared background image and the personalized concept images.  
This process yields edit-friendly latent representations and masks with accurately defined regions. 

In the second stage, we leverage the latents and masks obtained in the previous step to generate an image that seamlessly integrates the personalized concepts with the background. 
To accomplish this effectively, we introduce three core techniques. First, \textbf{Guided appearance attention} enhances appearance fidelity by reconstructing the keys and values of personalized concept images using corresponding target latents. Second, \textbf{Mask-guided noise mixing} selectively applies predicted noise to concept-specific regions, ensuring non-personalized areas are preserved. Third, \textbf{Background dilution} mitigates concept leakage by applying extended object masks during self-attention operations.
Together, these techniques enable FlipConcept to preserve concept identity, reduce overfitting, and improve compositionality in complex multi-object scenes—all without model retraining.  
Our method thus provides an efficient and modular approach for multi-concept personalization in diffusion models.  
Experimental results verify that FlipConcept generates coherent, high-quality images while preserving non-edited regions. Furthermore, the generated images achieve strong performance in CLIP~\cite{radford2021learning} and DINO~\cite{caron2021emerging,oquab2023dinov2} evaluations, highlighting the relevance and fidelity of the generated content.

\section{Related Works}

\subsection{Diffusion-based Text-to-Image Generation and Editing}
Denoising diffusion models~\cite{ho2020denoising,song2020denoising} and their extensions~\cite{rombach2022high,podell2023sdxl} have enabled high-quality image synthesis from text prompts by leveraging vision-language encoders such as CLIP~\cite{radford2021learning, lee1990translation}.
Recent approaches have extended these models to image editing by manipulating attention mechanisms within the denoising U-Net, allowing for localized control over appearance and structure~\cite{cao2023masactrl,nam2024dreammatcher,alaluf2024cross, lee2001automatic}. Recently, multi-indicator systems for image quality assessment~\cite{wang2025structural} have been proposed to improve the evaluation of synthesized images.

\subsection{Multi-Concept Personalization.}
Generating images that incorporate multiple user-specific concepts is an important yet challenging task. Techniques such as Textual Inversion~\cite{gal2022image} optimize text embeddings, while others either fully~\cite{ruiz2023dreambooth} or partially~\cite{kumari2023multi} fine-tune the model parameters, or apply rank-one modifications~\cite{tewel2023key}.
To address issues like concept blending or partial loss, recent approaches employ weight merging~\cite{gu2024mix} and training-free strategies (e.g., Concept Weaver~\cite{kwon2024concept} and DreamMatcher~\cite{nam2024dreammatcher}) during sampling, although some still require inversion or fine-tuning. Building upon these efforts, our method reduces additional optimization steps while enabling flexible and high-quality generation of multi-concept images. 

\section{Preliminary}
\subsection{Latent Diffusion Models}
Latent Diffusion Model (LDM)~\cite{rombach2022high} performs denoising in a compressed latent space for efficient and high-resolution synthesis. Input images are encoded into latent vectors and reconstructed via a decoder. To incorporate text conditions, we use a pretrained CLIP text encoder~\cite{radford2021learning} to generate a condition vector \(c\), guiding the generation to align with the prompt semantics. The model is trained to predict noise using the following loss function:
\begin{equation}
L = \mathbb{E}_{x_0, \epsilon \sim \mathcal{N}(0,I)}\Bigl[\bigl\|\epsilon - \epsilon_\theta(x_t, t, c)\bigr\|_2^2\Bigr],
\end{equation}
\noindent where \(x_t\) is the sample added noise in the \(t\)-th time step, and \(\epsilon_\theta\) is the diffusion model~\cite{ho2020denoising} with parameters \(\theta\). 

\subsection{Self-Attention in Stable Diffusion}

We adopt Stable Diffusion (SD)~\cite{rombach2022high}, an LDM-based model built on a denoising U-Net~\cite{ronneberger2015u} with self and cross-attention modules for processing latent representations. In particular, self-attention plays a critical role in refining the spatial layout and enhancing fine-grained details. At each time step \(t\), the noisy latent code \(z_t\) is transformed via intermediate features \(\phi_l(z_t)\), from which query, key, and value vectors are computed through learned linear projection:
\begin{equation}
\quad Q = W_Q\bigl(\phi_l(z_t)\bigr), \quad K = W_K\bigl(\phi_l(z_t)\bigr), \quad V = W_V\bigl(\phi_l(z_t)\bigr).
\end{equation}

These are used to compute attention scores via scaled dot-product attention, which captures the contextual relevance between spatial locations~\cite{LEE1995783}:
\begin{equation}
\mathrm{Attention}(Q, K, V) = \mathrm{softmax}\left( \frac{Q K^T}{\sqrt{d_k}} \right) V,
\end{equation}
\noindent
where \(d_k\) is the dimension of the key vectors.

\section{Method}

\begin{figure*}[htb!]
\centering
\includegraphics[width=0.7\textwidth]{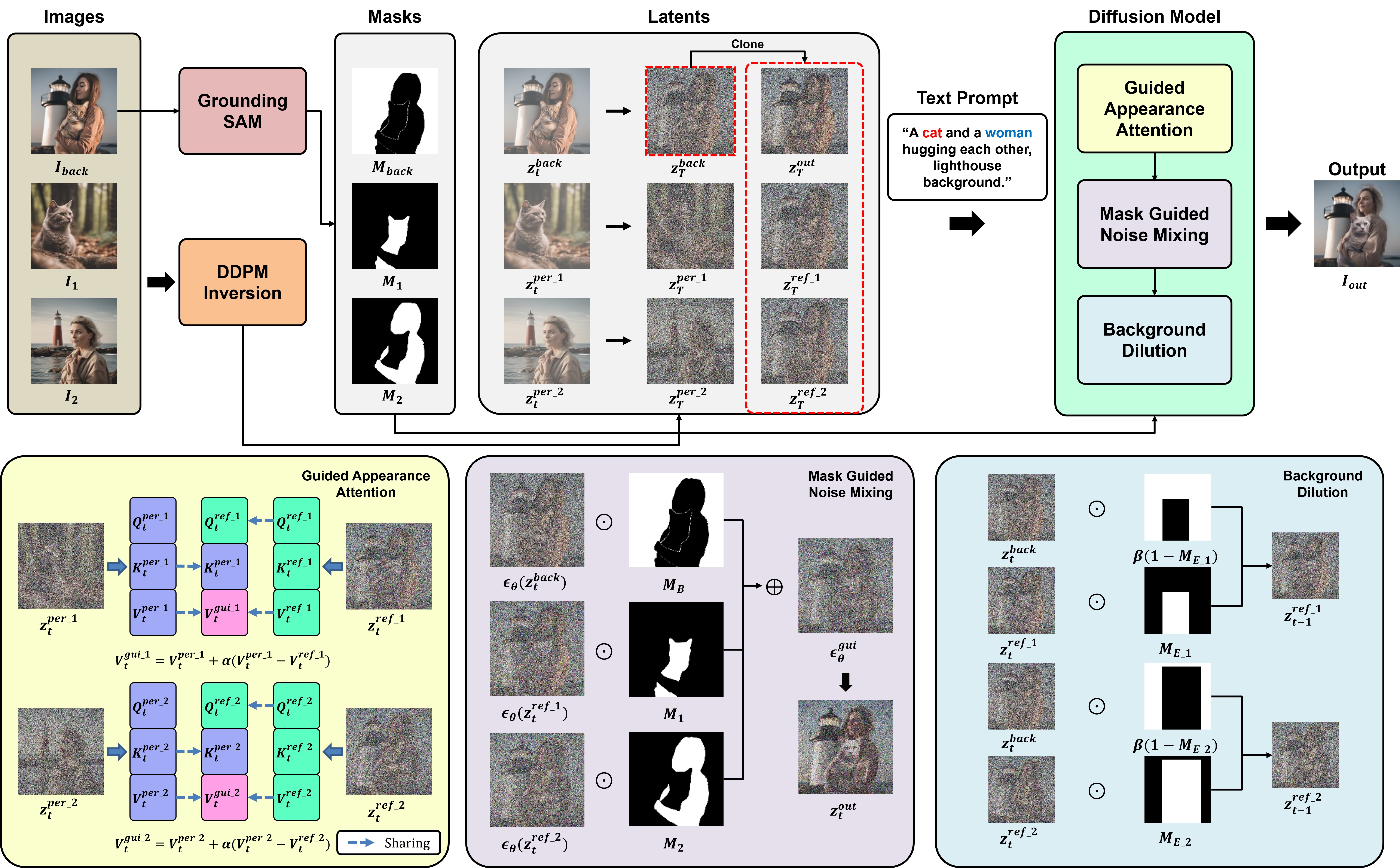}
\caption{\textbf{Overview of FlipConcept.}
Our framework leverages Grounded SAM~\cite{ren2024grounded} and DDPM Inversion~\cite{huberman2024edit} to extract masks and latents of both background and personalized concepts, which are then combined through adaptive concept blending including guided appearance attention, mask-guided noise mixing, and background dilution. Notably, in our method, $I_{\text{back}}$ can be either randomly generated or intentionally selected, while prior tuning-based methods typically rely on random background generation, offering limited control over user intent.} 
\label{fig}
\end{figure*}

In this section, we introduce \textbf{FlipConcept}, a tuning-free framework for integrating multiple personalized concepts into a single image using one model. 
Inspired by step-by-step generation~\cite{kwon2024concept}, our method consists of two stages.
In the first stage, we generate latent representations and masks, and in the second, we use them to guide the diffusion model in producing a background image $I_{\text{out}}$ that coherently reflects all concepts.
An overview is shown in Fig.~\ref{fig}, with details presented in the following sections.

\subsection{Input Preparation}

\subsubsection{Background Image and Mask Generation}

Most existing approaches~\cite{kumari2023multi, gal2022image, tewel2023key, gu2024mix} randomly generate background regions while excluding those defined by personalized concepts. 
In contrast, our method generates multiple backgrounds with Stable Diffusion XL~\cite{podell2023sdxl}, keeps only those with CLIP text–image similarity above 0.3~\cite{radford2021learning}, and lets users pick one where the target object is clearly visible.

Once the background image is prepared, we extract object and background masks using Grounded SAM~\cite{ren2024grounded}, which integrates Grounding DINO~\cite{liu2024grounding} for detection and SAM~\cite{kirillov2023segment} for segmentation~\cite{fujisawa1999information}. These masks localize objects and define background regions to guide generation. Our approach assumes that each reference image sufficiently represents the corresponding concept, and that the extracted masks accurately delineate the object regions. These assumptions are standard in recent personalization work~\cite{kwon2024concept, kumari2023multi}, and our experiments confirm their practicality for typical user concepts.

\subsubsection{Image Inversion}

We extract latent representations from the background and concept images using Edit-Friendly DDPM Inversion~\cite{huberman2024edit}, which directly samples a noise map that preserves structural and visual details, unlike traditional DDIM Inversion~\cite{song2020denoising}. Edit-Friendly DDPM Inversion forms a sequence via a probabilistic path directly defined from the original image $x_0$ (e.g., $I_{\text{back}}$, $I_1$, and $I_2$ in Fig.~\ref{fig}).
To achieve this, at each time step $t$ ($1 \le t \le T$), we independently sample noise $\tilde{\epsilon}_t \sim \mathcal{N}(0, I)$ and construct $x_t$ as follows:

\begin{equation}
x_t = \sqrt{\bar{\alpha}_t} \cdot x_0 + \sqrt{1 - \bar{\alpha}_t} \cdot \tilde{\epsilon}_t,
\end{equation}

\noindent where ${\bar{\alpha}_t}$ is the cumulative noise schedule parameter. Because \(\tilde{\epsilon}_t\) is sampled independently, adjacent noise vectors become negatively correlated, increasing the divergence between \(x_t\) and \(x_{t-1}\) and increasing the variance of the extracted noise map \((z_T, \dots, z_1)\). Next, we invert the auxiliary sequence \((x_1, \dots, x_T)\) to extract the noise map. For each time step \(t\), given \(\hat{\mu}_t(x_t)\) and \(\sigma_t\), the noise component \(z_t\) is computed as:

\begin{equation}
z_t = \frac{x_{t-1} - \hat{\mu}_t(x_t)}{\sigma_t}.
\end{equation}

This edit-friendly noise map not only retraces the generative path but also provides a latent representation that faithfully encodes the input image, allowing for flexible editing without significant loss of the original details. The background latent \(z_T^{\text{back}}\) is replicated to exceed the number of personalized concepts (e.g., \(z_T^{\text{per}\_1}\) and \(z_T^{\text{per}\_2}\) in Fig.~\ref{fig}) by one copy. One copy is used to initialize \(z_T^{\text{out}}\) for generating the final output image \(I_{\text{out}}\), while the remaining copies (denoted as \(z_T^{\text{ref}\_1}, z_T^{\text{ref}\_2}, \dots\)) serve as reference conditions to robustly embed specific objects or concepts in subsequent editing steps. Finally, these latent representations, along with precisely defined masks, are fed into the Stable Diffusion model~\cite{rombach2022high} to generate the final image \(I_{\text{out}}\), which seamlessly integrates multiple personalized concepts into a coherent scene.

\subsection{Adaptive Concept Blending}

Once all inputs are prepared, we pass them to the diffusion model along with three key techniques \textbf{Guided appearance attention}, \textbf{Mask-guided noise mixing}, and \textbf{Background dilution} to address challenges in multi-concept blending. These methods ensure structural stability and precise appearance transfer, enabling seamless concept fusion and high-quality image generation.

\subsubsection{Guided appearance attention} Our proposed attention mechanism emphasizes appearance information by sharing the key and value of the personalized concept image in the self-attention layer and adjusting the key and value of the reference latent. Specifically, it modifies the self-attention mechanism by replacing the key of the reference latent with that of the personalized concept image, and refines the value using a value guidance mechanism inspired by classifier-free guidance~\cite{ho2022classifier}, which highlights the distinctive appearance characteristics of the personalized concept. This process improves both text–image alignment and overall visual quality. The value guidance is defined as follows:
\begin{equation}
V_t^{\text{gui}\_{\mathit{i}}} = V_t^{\text{per}\_{\mathit{i}}} + \alpha \cdot \bigl(V_t^{\text{per}\_{\mathit{i}}} - V_t^{\text{ref}\_{\mathit{i}}}\bigr),
\end{equation}
\noindent where \(V_t^{\text{per}\_{\mathit{i}}}\) represents the value of the \(i\)-th personalized concept latent (\(z_t^{\text{per}\_{\mathit{i}}}\)) at time step $t$, \(V_t^{\text{ref}\_{\mathit{i}}}\) is the value of the \(i\)-th reference latent (\(z_t^{\text{ref}\_{\mathit{i}}}\)) at time step $t$, and \(\alpha\) is a hyperparameter controlling how strongly the unique appearance of the personalized concept is emphasized.
In this paper, we set \(\alpha=0.15\), which we empirically found to balance visual quality and alignment through extensive preliminary experiments. By adjusting this hyperparameter, one can further refine the emphasis on appearance features based on specific use cases or preferences.
The modified key and guided value are then fed into the self-attention operation~\cite{cao2023masactrl, alaluf2024cross}, which outputs the attention-refined latent \(z_t^{\text{ref}\_{\mathit{i}}}\). This refined latent is subsequently passed to the denoising phase.

\subsubsection{Mask-guided noise mixing} To effectively combine noise predictions, we employ a mask-guided mixing approach that blends the noise predicted by a diffusion model at the mask unit, ensuring that each personalized concept is only applied within its designated region. Namely, the object mask (\(M_i\)) corresponds to the area of the concept to be edited, while the background mask (\(M_B\)) covers the remaining preserved region, and the noise is combined using a Hadamard product (\(\odot\)) approach.
The process can be formulated as follows:
\begin{equation}
\epsilon_\theta^{\text{gui}} = \epsilon_\theta(z_t^{\text{back}}) \odot M_B + \sum_{i=1}^n \epsilon_{\theta}(z_t^{\text{ref}\_{\mathit{i}}}) \odot M_i,
\end{equation}
\noindent where \(\epsilon_\theta^{\text{gui}}\) denotes the guided noise used in the denoising process~\cite{ho2020denoising,huberman2024edit} to generate the final output \(z_t^{\text{out}}\) at time step \(t\), \(\epsilon_\theta\bigl(z_t^{\text{back}}\bigr)\) represents the noise for the background, and \(M_B\) is the background mask.
Additionally, $\epsilon_{\theta}(z_t^{\text{ref}\_{\mathit{i}}})$ and $M_i$ correspond to the \(i\)-th reference noise and its mask, respectively.
Simultaneously, the noise corresponding to the \(i\)-th reference object is also updated by re-synthesizing the area outside the object's region with the background noise, using the \(i\)-th object mask. 
Moreover, a swap guidance mechanism~\cite{alaluf2024cross} is employed to direct the denoising process towards the desired appearance, reducing distortions and enhancing realism. As a result, one can clearly separate object and background regions, while naturally fusing various concepts from the text prompt into a single image.

\subsubsection{Background dilution} This mechanism reduces the influence of the background in the reference latent generated during the diffusion process by leveraging object masks.
After generating the reference latent, a minimal rectangular mask is derived from the object mask and used to mix the background latent with the reference latent, scaled by a weight \(\beta\) to control the blend ratio and dilute the background influence. The formula for updating the latent is as follows:
\begin{equation}
z_{t-1}^{\text{ref}\_{\mathit{i}}} = z_t^{\text{back}} \odot \beta \bigl(1 - M_E\bigr) + z_t^{\text{ref}\_{\mathit{i}}} \odot M_E,
\end{equation}
\noindent where \(z_t^{\text{back}}\) represents the background latent, \(z_t^{\text{ref}\_{\mathit{i}}}\) denotes the reference latent at time step \(t\), and \(\beta\) is a coefficient that adjusts their relative influence.
Empirically, we found \(\beta=0.8\) to yield the best results through similar preliminary experiments, ensuring robustness across different scenarios.
Moreover, \(M_E\) is a rectangular mask expanded from the object mask, precisely covering only the region around the target object.
This approach clearly separates the background and object while representing only the minimal area around the object, allowing personalized concepts to be reflected more accurately. 

\section{Experiments}

\begin{figure*}[htb!]
\centering
\includegraphics[width=0.62\textwidth]{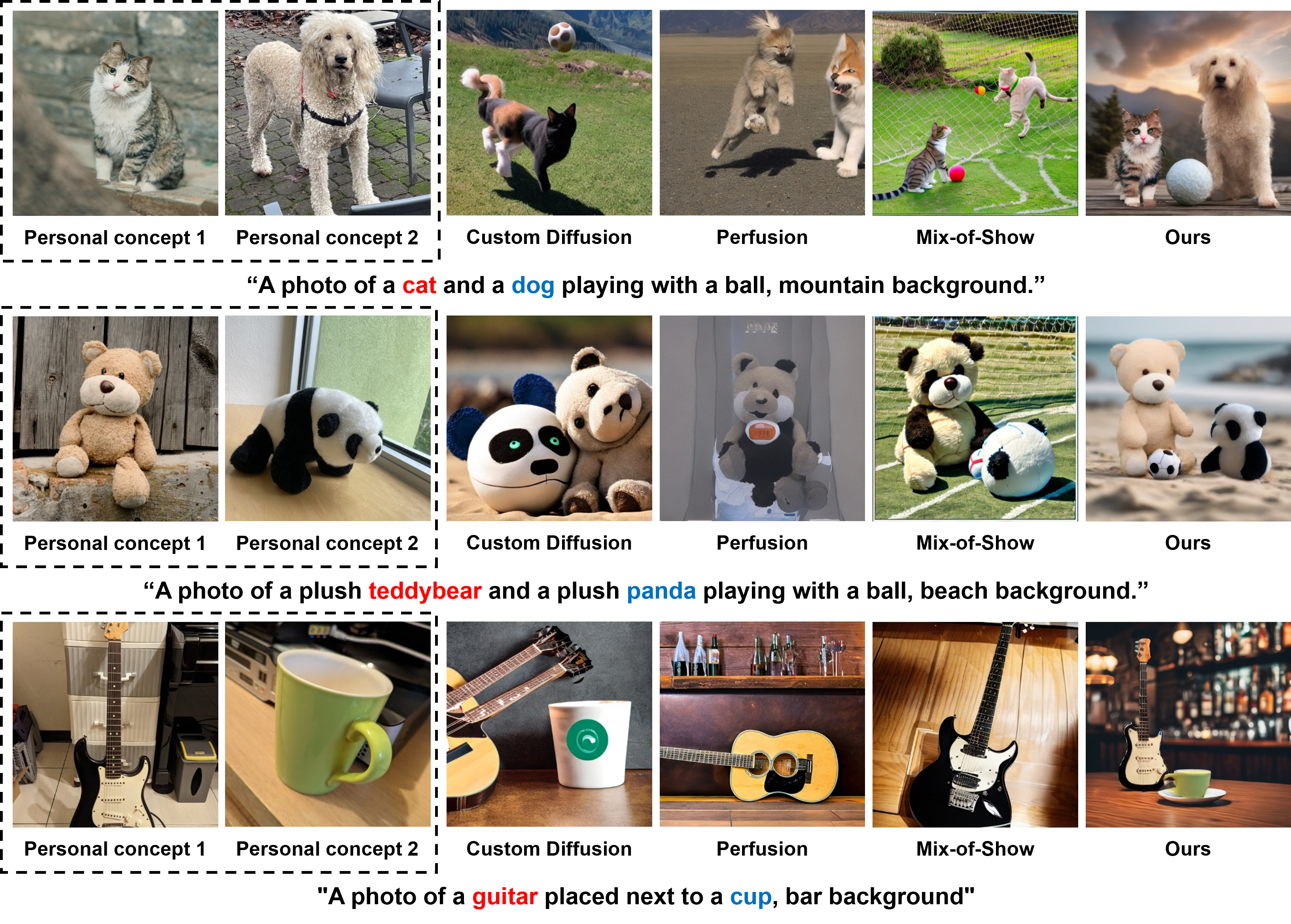}
\caption{\textbf{Qualitative Evaluation of Multi-Concept Personalization.} We compare our method FlipConcept against baseline approaches (Custom Diffusion~\cite{kumari2023multi}, Perfusion~\cite{tewel2023key}, Mix-of-Show~\cite{gu2024mix}) using prompts that incorporate multiple personalized concepts (e.g., cat and dog) with various backgrounds.
Our method generates coherent results in complex scenarios without undesired blending or loss of structural integrity.}
\label{fig3}
\end{figure*}

\begin{figure*}[htb!]
\centering
\includegraphics[width=0.72\textwidth]{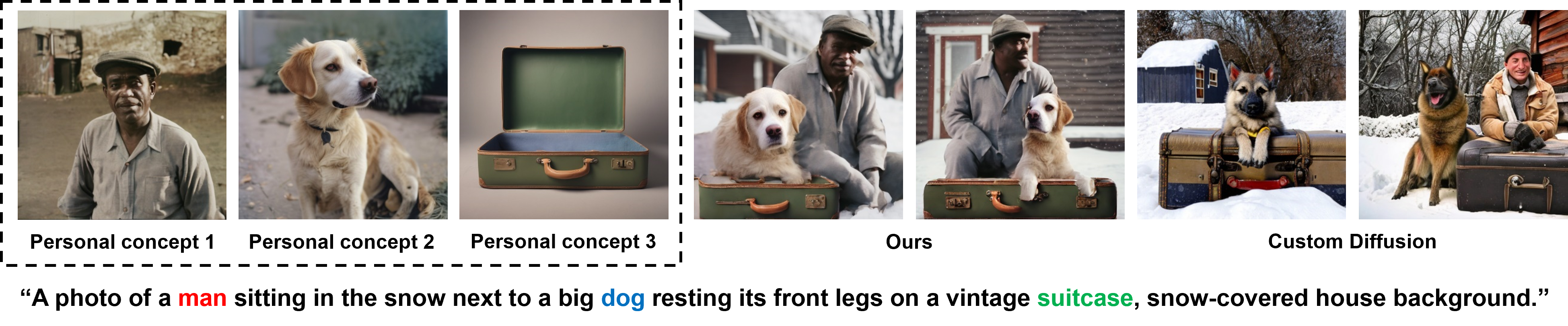}
\caption{\textbf{Robust Multi-Concept Personalization in Challenging Scenarios.} We compare our method against Custom Diffusion~\cite{kumari2023multi} on prompts involving three interacting concepts. Custom Diffusion~\cite{kumari2023multi} frequently omits or blends elements together, failing to preserve the intended scene composition. In contrast, our method preserves all target concepts and their spatial structure.}
\label{fig4}
\end{figure*}

\subsection{Experimental Settings}
We conducted all experiments using the Stable Diffusion v2.1 base model~\cite{rombach2022high}, generating images at a resolution of 512×512 pixels on two NVIDIA A40 GPUs. 
Following the evaluation protocols established in prior works on multi-personal concept personalization~\cite{kumari2023multi, tewel2023key, kwon2024concept}, we evaluated on the Custom Concept 101 dataset~\cite{kumari2023multi} and additional images generated by Stable Diffusion~\cite{rombach2022high}, using five evaluation sets with prompts automatically generated by ChatGPT~\cite{achiam2023gpt} to assess inter-object relationships.
For evaluation metrics, we adopted two CLIP-based metrics ($T_\text{CLIP}$ and $I_\text{CLIP}$)~\cite{radford2021learning} and one DINOv2-based metric ($I_\text{DINO}$)~\cite{caron2021emerging,oquab2023dinov2}. 
All metrics rely on cosine similarity. $T_\text{CLIP}$ evaluates the alignment between the generated image and the prompt, while $I_\text{CLIP}$ and $I_\text{DINO}$ measure concept fidelity based on region-specific embeddings.
We compared our method with Custom Diffusion~\cite{kumari2023multi}, Perfusion~\cite{tewel2023key}, and Mix-of-Show~\cite{gu2024mix}, demonstrating its superior quantitative, qualitative, and practical performance. 
Additionally, we conducted ablation studies to assess the contribution of each component of the framework.

\begin{table}[htbp]
\caption{\textbf{Quantitative comparison of multi-personalized concept generation.} Our method (d) outperforms the baseline approaches in all three metrics $T_\text{CLIP}$, $I_\text{CLIP}$ and $I_\text{DINO}$~\cite{caron2021emerging,oquab2023dinov2}, indicating superior text and concept alignment.}
\centering
\resizebox{0.7\linewidth}{!}{
\begin{tabular}{clccc}
\toprule
& \textbf{Method} & $T_\text{CLIP} \uparrow$ & $I_\text{CLIP} \uparrow$ & $I_\text{DINO} \uparrow$\\
\midrule
(a) & Custom Diffusion~\cite{kumari2023multi}          & 0.3480 & 0.8078 & 0.4657 \\
(b) & Perfusion~\cite{tewel2023key}                & 0.3044 & 0.7616 & 0.3267 \\
(c) & Mix-of-Show~\cite{gu2024mix}      & 0.3206 & 0.7920 & 0.4262 \\
(d) & \textbf{FlipConcept (Ours)}                           & \textbf{0.3810} & \textbf{0.8830} & \textbf{0.6502} \\
\bottomrule
\end{tabular}
}
\label{tab:component_comparison}
\end{table}

\subsection{Generation Results}

To evaluate the effectiveness of the proposed method, we generated 50 images per method and performed quantitative and qualitative comparisons.
Our results consistently demonstrate our method’s superiority in generating coherent and structurally accurate multi-concept images

Table~\ref{tab:component_comparison} summarizes the quantitative performance across CLIP-based~\cite{radford2021learning} and DINO-based~\cite{caron2021emerging,oquab2023dinov2} metrics. 
Our method achieves the highest scores in $T_\text{CLIP}$, $I_\text{CLIP}$, and $I_\text{DINO}$, indicating strong alignment between generated images and text prompts, as well as faithful representation of individual concept details. 
In contrast, baseline methods show noticeable performance degradation as the complexity (number of concepts) increases. 
For example, Custom Diffusion manages multiple concepts but exhibits structural inconsistency and blending, resulting in lower scores than ours.
 
To complement these quantitative results, Fig.~\ref{fig3} illustrates qualitative comparisons of multi-concept generation results across various challenging prompts. 
Our approach effectively handles complex interactions between personalized concepts, such as the prompt \textit{``A photo of a guitar placed next to a cup, bar background.''}
For example, Perfusion faces difficulties in maintaining the global image structure, whereas Mix-of-Show frequently blends adjacent concepts into ambiguous shapes. 
Conversely, our method integrates distinct concepts while preserving prompt fidelity and clarity.

When composing prompts with three interacting concepts, our method consistently outperforms Custom Diffusion~\cite{kumari2023multi}, as illustrated in Fig.~\ref{fig4}. Custom Diffusion~\cite{kumari2023multi} integrates multiple concepts either through joint fine-tuning or by merging separately fine-tuned models. However, as the number of concepts increases, this approach often leads to degraded image quality due to challenges in balancing and preserving each concept's representation. In contrast, our method maintains the integrity of each concept and preserves the intended structural relationships, even in complex scenarios.

\begin{table}[htbp]
\caption{\textbf{Total time comparison of multi-personalized concept generation.} FlipConcept achieves the best performance in total generation time, including additional training (such as fine-tuning or retraining) and inference.}
\centering
\resizebox{0.8\linewidth}{!}{%
\begin{tabular}{clccc}
\toprule
& \textbf{Method} & \textbf{Additional training} & \textbf{Inference} & \textbf{Total} \\
\midrule
(a) & Custom Diffusion~\cite{kumari2023multi} & 814s & 13s & 827s \\
(b) & Perfusion~\cite{tewel2023key} & 7115s & 7s & 7122s \\
(c) & Mix-of-Show~\cite{gu2024mix} & 1509s & 16s & 1525s \\
(d) & \textbf{FlipConcept (Ours)} & \textbf{0} & \textbf{86s} & \textbf{86s} \\
\bottomrule
\end{tabular}%
}
\label{tab:time_comparison}
\end{table}

Additionally, Table~\ref{tab:time_comparison} compares our approach to existing methods in terms of computational efficiency~\cite{ding2024freecustom}. Since our method does not require additional fine-tuning or retraining, it exhibits significantly improved time efficiency. 
Tuning-based models, by contrast, incur substantial additional training costs, limiting their practical applicability.
Our approach completely eliminates these additional training steps.
As a result, Table~\ref{tab:time_comparison} shows that our method has an additional training time of 0 seconds, underscoring the absence of any fine-tuning or retraining overhead.
Our method can thus be directly and efficiently deployed across diverse scenarios.

\begin{table}[htbp]
\caption{\textbf{Ablation Study.} Each component contributes meaningfully to our method's performance. The complete setting (d) achieves the best results across all metrics.}
\centering
\resizebox{0.9\linewidth}{!}{%
\begin{tabular}{clccc}
\toprule
& \textbf{Method} & $T_\text{CLIP} \uparrow$ & $I_\text{CLIP} \uparrow$ & $I_\text{DINO} \uparrow$\\
\midrule
(a) & Only mask-guided noise mixing            & 0.3737 & 0.8623 & 0.6005 \\
(b) & (a) + Guided appearance attention        & 0.3733 & 0.8711 & 0.6232 \\
(c) & (b) + Background dilution                & 0.3743 & 0.8655 & 0.6130 \\
(d) & \textbf{(c) + Mask-guided noise mixing on reference noise (Ours)} & \textbf{0.3810} & \textbf{0.8830} & \textbf{0.6502} \\
\bottomrule
\end{tabular}%
}
\label{tab:component_comparison2}
\end{table}

\subsection{Ablation Study}
Table~\ref{tab:component_comparison2} summarizes the results of an ablation study conducted to verify the contributions of each proposed component to overall method performance.

Specifically, setting (a), we apply only mask-guided noise mixing and key-value replacement \cite{cao2023masactrl, alaluf2024cross} within self-attention. 
Setting (b) introduces guided appearance attention, improving the clarity and preservation of target color and texture information. 
In setting (c), we apply background dilution, which attenuates irrelevant background details and further emphasizes the representation of the target concepts.
Finally, setting (d) applies mask-guided noise mixing directly to reference noise. This preserves fine details of the personalized concepts and reduces ambiguity among multiple targets, achieving the highest visual and textual similarity scores.

Overall, the ablation study quantitatively demonstrates that each component substantially enhances performance, confirming their effectiveness in multi-concept personalization tasks.

\section{Conclusion}
We presented FlipConcept, a novel tuning-free framework for integrating multiple personalized concepts into a single image. By leveraging guided appearance attention, mask-guided noise mixing, and background dilution, our method achieves structurally coherent and semantically faithful image generation without the need for additional fine-tuning.
Experiments validate its superior performance and efficiency over existing methods. Despite these advantages, our method may face limitations when objects are highly entangled or spatially overlapping, where mask-based separation can lead to slight visual distortion. Future work will explore more adaptive blending, such as soft mask adjustment for overlaps.

\end{document}